\documentclass[letterpaper, 10 pt, conference]{ieeeconf}  
\IEEEoverridecommandlockouts                              
\usepackage{graphicx}
\title{Processing Natural Language About Ongoing Actions}

\author{Steve Doubleday, Sean Trott, Jerome Feldman}

\begin{document}

\maketitle
\thispagestyle{plain}
\pagestyle{plain}

\begin{abstract}

Actions may not proceed as planned; they may be interrupted, resumed or overridden.  This  is a challenge to handle in a natural language understanding system.  We describe extensions to an existing implementation for the control of autonomous systems by natural language, to enable such systems to handle incoming language requests regarding  actions.  Language Communication with Autonomous Systems (LCAS) has been extended with support for X-nets, parameterized executable schemas representing actions.  X-nets enable the system to control actions at a desired level of granularity, while providing a mechanism for language requests to be processed asynchronously.  Standard  semantics supported include requests to stop, continue, or override the existing action.  The specific domain demonstrated is the control of motion of a simulated robot, but the approach is general, and could be applied to other domains. 

\end{abstract}

\section{Introduction}

A challenge for natural language understanding systems that interface with robots is responding to changes in the situation.  Actions may need to be interrupted, cancelled, stopped, resumed, or overridden.  Underlying these responses are assumptions about the state of events.  The temporal structure of events is expressed in linguistic constructions termed \textit{aspect} \cite{Narayanan1997}; an action can be about to start, or ongoing, or just completed.  An action that is interrupted or stopped is presumed to be ongoing; one that is cancelled is presumed not to have started yet.

We have extended an existing system, Language Communication with Autonomous Systems (LCAS) \cite{Trott2015}, to enable it to handle language requests regarding the actions of simulated robotic agents.  The extensions center around the use of \textit{X-nets} \cite{Narayanan1997}, which implement parameterized executable schemas representing actions.  An X-net is implemented as an extension to a  \textit{Petri net} \cite{murata1989petri}.  The Petri net formalism is useful for the description and analysis of concurrent processing in distributed systems.  An X-net implements many of the standard semantics we associate with action.  Actions can be enabled, started, ongoing, or done, which expresses the normal execution path.  Actions can also be suspended, resumed, or restarted, and handle exception conditions.  In addition to having a shared standard action semantics, each X-net is tailored to meet the demands of a set of related tasks.  In this case, a Move X-net serves as an interface to drive the motion of a simulated robot.  Asynchronous communication links the language side of the system, handling new user requests as they arrive, to the action side of the system, which drives the motion of the simulated robot.        

\section{X-Nets and Simulation Semantics}
There is considerable evidence that language understanding proceeds in part by simulating the actions implied by the text \cite{bergen2004simulated,bergen2012louder}. 
Actions and the simulation of actions can be modeled by executable schemas parameterized by language \cite{Narayanan1997,Narayanan2010}.  For example, verbs concerning motion can be organized in terms of an implicit ascending speed parameter:  ``crawl'', ``amble'', ``walk'', ``run'', ``dash''.  Linguistic aspect involves the use of grammar to parameterize the state of actions in reference to a general event model.  An action can be impending, under way, or completed.

\begin{figure}[!hbt]
\begin{center}
\includegraphics[scale=0.35]{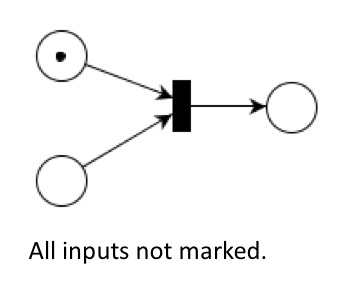}
\caption{Transition not yet enabled, as all input places are not marked.}
\label{fig:disabled}
\end{center}
\end{figure}

\begin{figure}
\begin{center}
\includegraphics[scale=0.25]{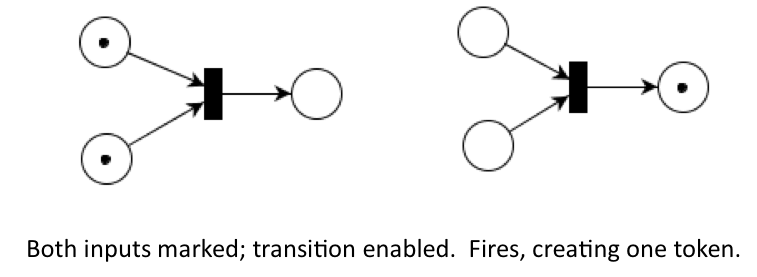}
\caption{When all input places are marked, transition is enabled and fires, marking the output place}
\label{fig:enabled}
\end{center}
\end{figure}

An executable schema can be implemented as an X-net, a parameterized Petri net representation of an action. The Petri net formalism models distributed processing as a set of states and transitions among states, connected by directed arcs. Transitions fire when the pre-conditions defined by their input arcs are met, generating updates to the places connected to their output arcs.  The pre- and post-conditions take the form of one or more tokens in a place.  The vector of token counts for all places in the Petri net constitutes its current state, or \textit{marking}.   Figure~\ref{fig:disabled} depicts a transition that will not fire; Figure~\ref{fig:enabled} depicts an enabled transition before and after firing; after firing the transition is disabled.  In this case, firing consumes two tokens and creates one token.


\begin{figure}
\begin{center}
\includegraphics[scale=0.4]{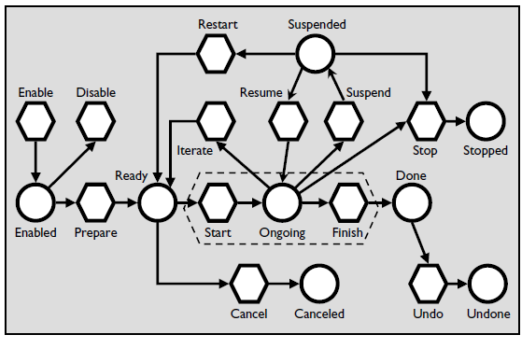}
\caption{Event graph (controller) depicting the structure of events \cite{Narayanan2010}}
\label{fig:controller}
\end{center}
\end{figure}

X-nets define a standard action semantics, states and transitions that are common to all actions. Depicted as an event graph in Figure~\ref{fig:controller}, this serves as a useful intermediate-level abstraction for the representation of actions \cite{Narayanan2010,doubleday2017computational}.  Scalar parameters can be passed to an X-net, enabling it to perform computations to modify its processing, e.g., to move faster or slower.  X-nets can also be composed of more granular X-nets, where each X-net has some ability to perform error recovery locally, with escalation to higher levels as needed.  

These X-net capabilities enable a system to control actions.  They also enable a system to reason in a consistent way about actions.   X-nets provide a natural way to model linguistic aspect.  By examining the current marking of an X-net, we can distinguish between actions that are about to begin, or are ongoing, or just completed.  X-nets can also support reasoning about events in general \cite{sinha2008answering}, including hypothetical or counter-factual language.

\section{System Architecture}


\subsection{System for Natural Language Understanding}

\begin{figure}[!hbt]
\includegraphics[width=\columnwidth]{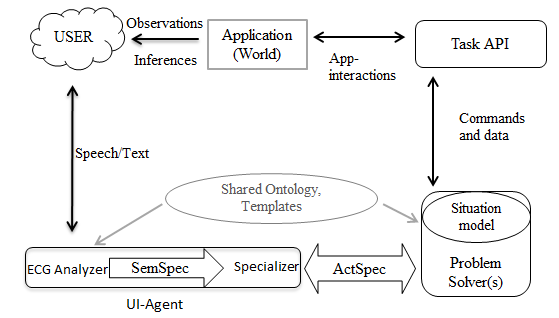}
\caption{System Architecture for natural language understanding.}
\label{fig:nlu_system}
\vspace{-10pt}
\end{figure}

We have implemented a system for natural language understanding \cite{Trott2015, Khayrallah2015,Eppe2016b}. The general architecture for this system is depicted in Figure \ref{fig:nlu_system}. Most of our previous work has focused on extending the core system framework to the robotics domain, in applications such as Morse \cite{Echeverria2011} and ROS \cite{Eppe2016}, but the system was designed to facilitate simple re-targeting to new domains and applications. Ongoing work includes applications such as real-time strategy games (StarCraft), metaphor analysis, and mental space modeling.

The system is modular; at the highest level, it can be divided into a \textit{language side} and an \textit{action side}. The language side, depicted on the left in Figure \ref{fig:nlu_system}, receives speech or text as input, and produces a structured representation of the text's meaning called an \textit{Action Specification} (ActSpec). This ActSpec, which can describe a command, query, or assertion about the world, is communicated as a JSON object to the action side of the system. The action side, depicted on the right in Figure \ref{fig:nlu_system}, consists of a \textit{Problem Solver}, a \textit{world model}, and a \textit{Task API}.  The Problem Solver unpacks the ActSpec to determine what task is to be accomplished and with what parameters.  The Problem Solver refers to the world model to determine the constraints on the possible solutions to the problems posed by the target task, and computes a solution.  Finally, the Problem Solver makes API calls to the underlying application, which constitute the actions necessary to accomplish the target task. Communication between the language and action sides is enabled by shared ActSpec templates, which guide ActSpec creation, and a shared ontology. The separation of the language and action sides allows them to be run on separate processes; thus, the action-side can solve ActSpecs while the language-side continues to receive text and speech input.

A crucial theoretical underpinning of the system design is previous work on mental simulation of language \cite{bergen2012louder}. The language side produces the parameters for the simulation of an event, which can then be simulated or executed by the action side.

More concretely, the ECG Analyzer \cite{Bryant2008}, a cognitively plausible language parser, uses an \textit{Embodied Construction Grammar} \cite{bergen2004simulated,Feldman2009} to produce a \textit{Semantic Specification} (SemSpec) of the events described in language. The SemSpec maps constructional information to the meaning of the sentence. Task-relevant information from the SemSpec is extracted by the Specializer and formatted into an ActSpec. This ActSpec provides the parameters for the simulation of language.

In the system described in \cite{Khayrallah2015,Trott2015}, the action side executed these incoming ActSpecs as they arrived. In the current work, X-nets are introduced to provide additional mechanisms for simulation and control.


\subsection{Integration of X-nets}

 The architecture depicted in Figure \ref{fig:nlu_system} has been modified to have the Problem Solver execute one or more X-nets to represent the action being undertaken through the task API. The Problem Solver examines the state of the X-net either during or after execution to make inferences about progress on the task.  Integrating X-nets into the architecture enables the system to deal with the following types of tasks, events, or requests:

\begin{itemize}
\item suspension: ``Robot1, stop moving!''
\item resumption:  ``Robot1, continue moving!''
\item interruption and redirection: while a move is in progress, e.g. ``Robot1, move to the green box!'', a new target is specified:   ``Robot1, move to the blue box!''
\end{itemize}

For the application of controlling a simulated robot, the Problem Solver invokes the Move X-net (Figure~\ref{fig:move-xnet}).  The Move X-net supports an additional parameterization, implementing different speeds for different verbs:  amble (slow), move (normal), dash (fast).

\subsection{X-net Problem Solver}

The incorporation of X-nets into the LCAS architecture allows a systematic decoupling of the language side and action side of the architecture, such that the user can interact without restriction with the system, often in response to the current system behavior.  The user can interrupt, cancel, resume, or restart the action, as appropriate, without being constrained by blocking or synchronous control flows.  As is seen in both biological and mechanical action loops, there is some inherent delay between a new request and the action response, but the asynchronous nature of Petri net execution enables this to be handled in a principled way, without resort to complex logic or ad hoc exception handling.  

To support this decoupling, the execution of an X-net Problem Solver involves two threads, one that listens on a queue for new incoming requests from the language side, and a second thread that interacts with the particular X-net appropriate to the task at hand.  In turn, the X-net interacts with the task API to accomplish the objective.  For an ongoing action, a new incoming request will be translated by the Problem Solver into an update to one or more places in the currently executing X-net, which in turn will cause a change in the X-net flow of control, corresponding to the meaning of the new request.  Requests regarding actions that have not yet started or have completed may result in the invocation of a new X-net, or no change.     

\subsection{Petri net Extensions to Support External Systems}

X-net support is built on PIPE V5, an open source Java Petri net editor and debugging environment\footnote{https://github.com/sarahtattersall/PIPE} \cite{bonet2007pipe,dingle2009pipe2,tattersall2014pipe}.  To support the current work, PIPE was extended to provide support for interfaces to and from external systems, and for execution through an API\footnote{Current code is here: https://github.com/sjdayday/PIPECore/tree/hierarchical-nets, and is documented here: https://github.com/sjdayday/PIPECore/wiki.  At some point, the extensions will be integrated as a release into PIPE V5}.  The extensions include:

\begin{itemize}
\item External Transition:  an extension to a Petri net transition, providing a mechanism to execute Java code whenever the transition fires.  The External Transition has access to the state of the Petri net, and is optionally given a context passed from the external system, consisting of an instance of an arbitrary Java class. 
\item External Input Place:  an extension to a Petri net place, providing a mechanism for an external system to update the marking of the place with one or more tokens.
\item External Output Place:  an extension to a Petri net place, providing a mechanism for an external system to be notified whenever the marking of the place changes.
\item PNML extensions:  extensions to the Petri Net Markup Language, which defines the specification of a Petri net in XML format, to support external transitions and places.
\item Merge Place:  an extension to a Petri net place, enabling a Petri net to be composed of multiple other Petri nets for purposes of design and coding.  Merge places serve to connect the various Petri nets together, and are then collapsed to single places at execution time, creating a single executable Petri net.
\item Runner: the Runner interface enables an external system to interact with a Petri net.  The interface supports the following functions:  
\begin{itemize}  
\item load a Petri net from a PNML file
\item start Petri net execution
\item mark a place in a Petri net
\item subscribe to notifications of a change in marking of a place in a Petri net
\item pass an object to be made available to a Petri net transition when it fires
\item subscribe to notifications of global Petri net events, including starting or stopping execution, and firing a transition
\end{itemize}
\end{itemize}

\subsection{Standard Action Semantics and the Move X-net}

\begin{figure}[!hbt]
\begin{center}
\includegraphics[scale=0.35]{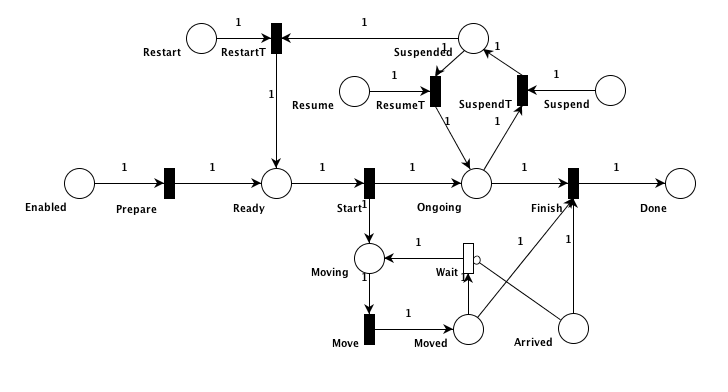}
\caption{Move X-net}
\label{fig:move-xnet}
\end{center}
\end{figure}

Figure~\ref{fig:move-xnet} shows the Move X-net, which controls the motion of a simulated robot.  The upper portion of the X-net is an implementation of a subset of the standard action semantics depicted in Figure~\ref{fig:controller}. Updating or reading places in this X-net provides a simple, standard interface to many of the operations that we assume should be available in any action -- starting, stopping,  resuming or restarting an action.  

The lower portion of the Move X-net depicts the logic specific to controlling motion.  This is a simplified interface that defers specifics of trajectory planning to the Problem Solver and to the Morse simulator, but additional logic could be added as needed for other domains, without affecting the operation of the standard action semantics in the upper portion of the X-net.   

\subsection{X-net Interface to an External System}

\begin{figure}[!hbt]
\begin{center}
\includegraphics[scale=0.35]{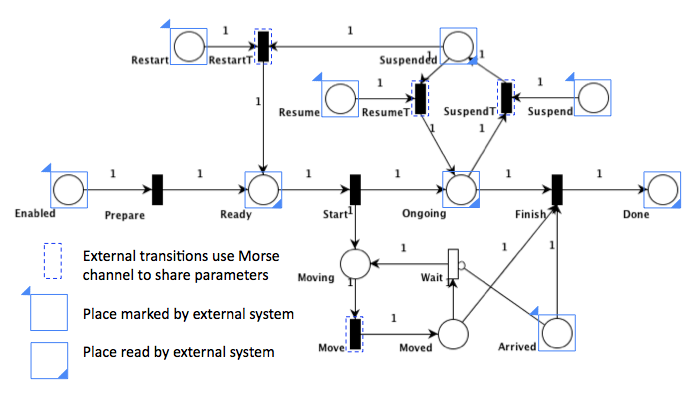}
\caption{Move X-net interface to an external system}
\label{fig:external-interfaces}
\end{center}
\end{figure}

The mechanics of the interface between the LCAS system and the Petri net implementation of the Move X-net is depicted in Figure~\ref{fig:external-interfaces}.  This serves as an example of the general interaction between an X-net and any external system.  

From the perspective of the X-net, the external system here is the Problem Solver component of the LCAS system, which controls execution of the motion by populating some subset of the standard external input places:  Enabled, Suspend, Resume and Restart.  The Problem Solver monitors progress of the motion by listening for updates to standard external output places:  Ready, Ongoing, Done, and Suspended.  The  interaction with the task API, in this case the Morse simulator,  happens through bi-directional updates to an instance of the task-specific MorseChannel Java class.  These updates happen during the external transitions: Move, SuspendT, ResumeT, and RestartT.  Each transition communicates an appropriate target operation, as well as the target position of the simulated robot, from the perspective of the X-net.  When any transition fires, control passes from the X-net back to the Problem Solver, which performs any appropriate pending action and  updates the current position of the robot from the perspective of the Morse simulator, and then returns control back to the X-net.  Finally, when the Morse simulator detects that the simulated robot has arrived at the destination, it updates the Arrived external output place to trigger transition of the Move X-net to the Done state. 

The mechanics of the integration between the Python implementation of LCAS and the Java implementation of the Petri net is handled through the JPype Python package.  JPype enables a Python script to start and attach a new JVM to the current Python thread, and then to create instances of Java classes.  The methods and public fields of the Java instances are then available for inspection and update by the script through syntax that is consistent with Python.  The first Problem Solver script creates the JVM, and any subsequent Problem Solver instances use the same JVM.  Each Problem Solver then creates an X-net as needed for its current task, using the Runner API, and interacts with that X-net until its task is completed.  

The interaction between the Problem Solver and the X-net can be configured to be effectively synchronous, such that every state change in the X-net enables the Problem Solver to regain control.  Depending on the task, however, the Problem Solver may not need such granular control, and may only register for asynchronous notifications.  Regardless of the degree of coupling between the Problem Solver and the X-net, this does not alter the overall asynchronous relationship between the language side and the action side in the LCAS architecture.  The processing of requests from the end user proceeds asynchronously from the combined activity of the Problem Solver, X-net, and supporting task API.  The standard action semantics provide natural support for the interruption of ongoing action, the preemption of pending action, or no operation when the action is already complete.  

\section{Robot Motion Example}

\subsection{Normal Motion}

\begin{figure}[!hbt]
\begin{center}
\includegraphics[scale=0.35]{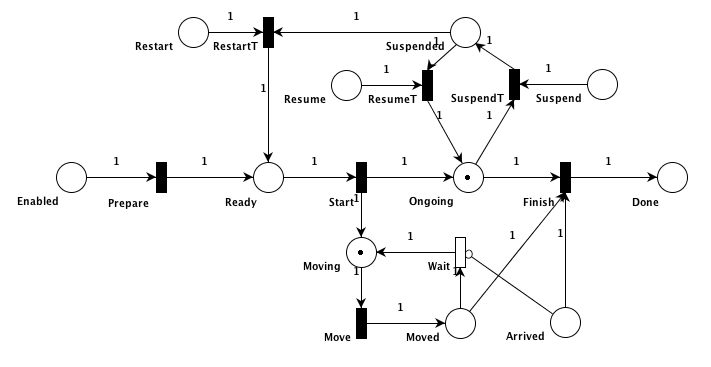}
\caption{Motion in progress}
\label{fig:move-ongoing}
\end{center}
\end{figure}

The integration of the Problem Solver, Move X-net, and the Morse simulator to accomplish motion of the simulated robot, has been demonstrated in the context of various language commands\footnote{Demonstration video: https://youtu.be/8PwHpng3Nj8}.  In response to the command ``Robot1, move to the blue box!'', the language side generates an ActSpec to the action side, adding it to the queue of requests to be processed by the X-net Morse Problem Solver.  The Problem Solver listens for each firing event in an executing X-net, which gives it an opportunity to check the queue for any new incoming requests.  This ensures that latency involved in processing a new request is short.  In this case, the incoming request causes a new instance of a PetriNetRunner to be created, which implements the Runner API.  Next, an instance of the Move X-net is instantiated through the Runner API, and the Problem Solver subscribes for updates to the various external output places of interest in the X-net, primarily those defined in the standard action semantics.  A new instance of a MorseChannel is created, to act as a shared context for all of the External Transitions in the Move X-net, and as the communication channel between the X-net and Morse.  The Problem Solver updates the MorseChannel with the location of the blue box, and the default speed.  The Problem Solver requests that the Enabled place in the X-net be updated with a single token, and then requests that the X-net be run.  The PetriNetRunner starts the execution of the Petri net, and notifies the listening Problem Solver of every event to which it has subscribed.  As noted earlier, these notifications include the firing of every transition, so that any new incoming language requests can be processed promptly.  Flow of control passes through the Ready place and then through the Start transition, causing both the Ongoing and Moving places to be populated, as depicted in Figure~\ref{fig:move-ongoing}.  

When the Move External Transition fires, the Morse Channel is updated with the target operation of ``move''.  The target location having been previously populated by the Problem Solver, control passes back to the Problem Solver, which passes the MorseChannel information to the Morse simulator, which begins the simulated robot motion.  Control then returns to the X-net, which loops through the Moved place and the Wait transition, before returning to Moving.  

The Wait transition populates the Moving place.  The transition is designated as a timed transition, to wait a configurable amount of time before firing.  Currently, however, the transition fires immediately, pending addition of the necessary support to the underlying PIPE implementation.  Implementation as a timed transition is intended as a performance optimization, to prevent the Moving / Moved loop from consuming system resources unproductively, and as a better model of the delays involved in physical motion. 

Each time the MorseChannel information is passed to Morse, Morse updates the current location of the robot; this information is currently unused by the X-net, but would be available if more advanced functions were required.  Finally, when the simulated robot reaches the target location, the MorseChannel is updated, and the Arrived place is marked, driving the X-net through the Finish transition to the Done place.  The Problem Solver is notified that the X-net is Done, indicating normal completion of the motion. 

\subsection{Processing Other Requests}

The power of the standard action semantics is seen when the system processes other incoming requests.  Requests of very different syntactic form and meaning are handled with little additional complexity.  ``Robot1, stop moving!'' causes the Problem Solver to populate the Suspend place.  If the X-net action is currently Ongoing, this causes the SuspendT External Transition to fire.  This transition  updates the MorseChannel target operation to be ``suspend'', which in turn causes Morse to stop the motion of the robot simulator.  The transition's firing also updates the state of the X-net to Suspended, causing the Problem Solver to be notified that motion has stopped.  

In response to the request ``Robot1, continue moving!'', the Problem Solver populates the Resume place.  If the current state of the X-net is Suspended, this drives the ResumeT External Transition, which updates the MorseChannel target operation to ``resume''.   This causes Morse to start the simulated robot moving again, resuming its trajectory toward the current goal.  Note that race conditions are handled simply; if the X-net is not in a Suspended state, the update of the Resume place has no impact on processing; it is effectively ignored.  

\begin{figure}[!hbt]
\begin{center}
\includegraphics[scale=0.35]{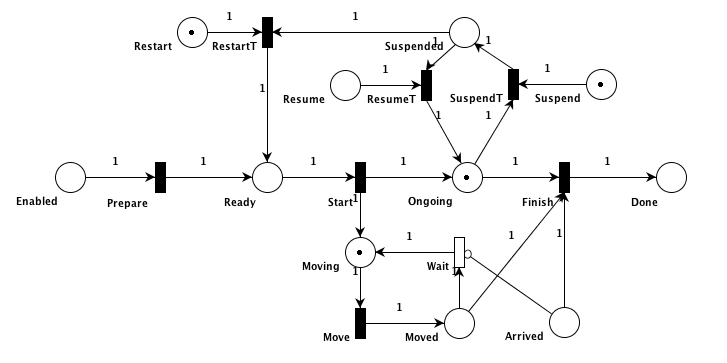}
\caption{Motion to be interrupted and re-directed}
\label{fig:move-restarting}
\end{center}
\end{figure}

Finally, the more complicated case of interruption and re-direction is also handled simply.  The user might wish to override a request that is in progress, e.g., ``Robot1, move to the blue box!''  The new request might be ``Robot1, dash to the green box!''  This new request causes the Problem Solver to calculate a  trajectory towards the green box, and to update the MorseChannel both with the location of the green box, and with a higher speed value.  As well, both the Suspend and Restart places are marked as in Figure~\ref{fig:move-restarting}.   X-net execution first transitions to the Suspended state, as in the ``stop moving!'' example, causing Morse to stop motion of the simulated robot.  The population of both the Restart and the Suspended places then drives the firing of the RestartT External Transition, which updates the MorseChannel target operation to ``restart''.  This causes Morse to start the motion of the simulated robot again, following the new trajectory and speed.  The X-net flow of control then passes back through the Ready place and the Start transition, re-entering the Ongoing state.       

\subsection{Non-Linguistic Events}\label{sec:nonlinguistic_interrupts}

In addition to linguistic requests, a robot frequently encounters other events as it solves various tasks. Many of these events involve changes to the environment, such as a moving object suddenly crossing paths with the robot's trajectory, or encountering a previously unknown obstacle. Environmental interrupts are challenging because the robot must dynamically adjust its world model to reflect these changes, and then develop a new plan to continue solving the original task. 

We implemented a general protocol for how the Problem Solver should behave when experiencing world-based interrupts, and extended this to the Morse domain. In the Morse demo \cite{Trott2015}, we addressed the problem of encountering objects that were not previously in the Problem Solver's world model. Novel objects could potentially disrupt the Problem Solver's original plan for solving a task; for example, if the robot is asked to a push a box north, it might find that a previously unknown box is in the way of its original path.

On the Morse side, our solution involved attaching a simulated proximity sensor\footnote{http://www.openrobots.org/morse/doc/1.2/user/sensors/proximity.html} to the instance of the robot model. Whenever the robot passes within a certain threshold of distance from an object, the proximity sensor relays information back to the Problem Solver about the object's location, color, and size. If the Problem Solver already knows about the object, the world model is simply verified to make sure the object location is correct. If the object is unknown, however, the Problem Solver performs three functions, in order:

\begin{enumerate}
\item{The Problem Solver updates its world model with the new information.}
\item{The Problem Solver sends an ActSpec back to the UI-Agent, to notify the human user that a new object has been discovered.}
\item{The Problem Solver develops a new plan for the original task, now taking into account the updated  world model.}
\end{enumerate}

Previous work also addressed the problem of disparate world models among multiple agents \cite{Trott2015}; in that case, (2) also involves sending notification ActSpecs to the other robots, so that they can also update their world models.



\section{Conclusion}

We have demonstrated the ability to handle language requests regarding the current state of actions in a graceful fashion, enabling actions to be interrupted, resumed and overridden.  This is a useful advance in the natural language control of autonomous systems.  The separation of language inputs from the control of actions enables both to proceed asynchronously.  This enables the overall system to be both responsive to new requests, while maintaining control of actions at an appropriate level of granularity.  Although demonstrated in the limited domain of movement of a simulated robot, the approach can be extended to other domains in a straightforward manner.  

\subsection{Limitations}

There are several limitations of the system.  Support in PIPE for external interfaces is limited to the API described earlier; interaction through a graphical interface is not yet supported, but should be available in the near future.  The standard semantics of X-nets invites some programmatic support; this has been discussed\footnote{https://github.com/sjdayday/xschema/wiki} but work has not yet begun.  Finally, no effort has been made to optimize the system for performance and scale.  

\subsection{Future Work}

The system supports current action.  A next step is to have the system support simulated action, where the execution of X-nets can be used to reason about the consequences of possible future actions \cite{sinha2008answering}.  This would enable the evaluation of counterfactual or hypothetical statements, such as 
``Robot1, if you moved north of the green box, could you push it south?''

Future work could also examine the problem of integrating our work on non-linguistic events (see Section \ref{sec:nonlinguistic_interrupts}) with the mechanism for X-net control.



\section*{Acknowledgements}
We thank Luca Gilardi for his help in designing the system framework, and for the creation of the GUI for the ECG Workbench Editor.  Professor William Knottenbelt and his students at Imperial College, London, created and have maintained PIPE through many releases.  Sarah Tattersall performed a thorough refactoring of the PIPE code base, without which the external system extensions would not have been possible.

\bibliographystyle{plain}
\bibliography{library}

\end{document}